\pdfoutput=1
\documentclass[11pt,a4paper]{article}
\usepackage[hyperindex,breaklinks]{hyperref}
\usepackage[hyperref]{acl2021}
\usepackage{times}
\usepackage{latexsym}

\newtheorem{hypothesis}{Hypothesis}
\usepackage{enumitem}
\usepackage{float}
\usepackage{graphicx}
\usepackage{multirow}

\usepackage{color}

\usepackage{microtype}

\aclfinalcopy 

\title{UoB at SemEval-2021 Task 5: Extending Pre-Trained Language Models to Include Task and Domain-Specific Information for Toxic Span Prediction}

\author{Erik Yan\\
  School of Computer Science\\
  University of Birmingham\\
  United Kingdom\\
  \texttt{\small erikdyan@outlook.com}\\\And
  Harish Tayyar Madabushi\\
  School of Computer Science\\
  University of Birmingham\\
  United Kingdom\\
  \texttt{\small Harish@HarishTayyarMadabushi.com}\\}

\date{}

\begin{document}
\maketitle
\begin{abstract}
Toxicity is pervasive in social media and poses a major threat to the health of online communities. The recent introduction of pre-trained language models, which have achieved state-of-the-art results in many NLP tasks, has transformed the way in which we approach natural language processing. However, the inherent nature of pre-training means that they are unlikely to capture task-specific statistical information or learn domain-specific knowledge. Additionally, most implementations of these models typically do not employ conditional random fields, a method for simultaneous token classification. We show that these modifications can improve model performance on the Toxic Spans Detection task at SemEval-2021 to achieve a score within 4 percentage points of the top performing team.
\end{abstract}

\section{Introduction and Motivation}
\label{sec:introduction-and-motivation}

Moderation is crucial to promoting healthy online discussions. The anonymity afforded by computer-mediated communication enables individuals to engage in toxic behaviour which they would otherwise not consider. Although many datasets and models focusing on toxicity detection have been released, most of them classify entire sequences of text, and do not highlight the individual words that make a text toxic. The Toxic Spans Detection task at SemEval-2021 \cite{pav2020semeval} focuses on the evaluation of systems that can accurately identify toxic spans within text. Highlighting such spans can provide more information to human moderators in the form of attribution, instead of an unexplained toxicity score per post, and is thus a crucial step towards successful semi-automated moderation. In this paper we focus on the shared task, wherein systems are expected to extract a list of toxic spans, or an empty list, per text. A toxic span is defined as a sequence of words that contributes to a text's toxicity.

Since 2018, NLP models have adopted the concept of generative pre-training on a diverse corpus of unlabelled text, followed by supervised fine-tuning on specific tasks \cite{radford2018language}. Pre-trained models are built to simulate anthropomorphic learning, wherein existing knowledge can be adapted to new tasks without the need to train on these tasks from scratch - a requirement of traditional machine learning models. This idea of transfer learning, whilst powerful, leads to models being fine-tuned on target tasks using significantly fewer epochs than was previously standard. This reduced training on the target task means that task-specific statistical information or domain-specific knowledge may not be learned by these models.

Such task-specific data may include count-based information, which has been shown to improve the performance of pre-trained models in sequence classification tasks \cite{lim2020uob,prakash2020incorporating}, or domain-specific knowledge, such as information pertaining to word toxicity, which has been shown to be one of the most predictive features of offensive commentary \cite{noever2018machine}.

Additionally, pre-trained models tend to use a fully connected layer for classification tasks. This classification layer, however, makes an individual localised prediction for each token without accounting for predictions made on other tokens. A CRF \cite{crf}, on the other hand, maximises the probability of the entire sequence of predictions. This makes it more effective for cases where neighbouring predictions may influence each other. NER is one such application, where the decision to assign a certain label to a token may be influenced by the labels assigned to neighbouring tokens. \citet{souza2020portuguese} combined the transfer capabilities of BERT \cite{devlin2019bert} with the structured predictions of a CRF, with the addition of a CRF yielding performance improvements in several token-level tasks.

Thus, this work aims to test the following hypotheses:

\begin{hypothesis}
    Count-based information can aid pre-trained models in token classification tasks.
\end{hypothesis}

\begin{hypothesis}
    Pre-trained models are unlikely to capture domain-specific information. Such information is likely to improve their performance in token classification tasks.
\end{hypothesis}

\begin{hypothesis}
    Adding a CRF, which affords a sentence-level predictive scope, will improve pre-trained model performance in token classification tasks.
\end{hypothesis}

To ensure reproducibility, our program code, including hyperparameters, is made available online\footnote{\url{https://github.com/erikdyan/toxic_span_detection}}.

\section{Related Work}
\label{sec:related-work}

Count-based information has been shown to improve the performance of pre-trained models in sequence classification tasks. \citet{lim2020uob} proposed an ensemble model of BERT and TF-IDF, which combined the sentence-level information captured by BERT with the corpus-level information provided by TF-IDF. The ensemble model performed 5 percentage points better than a standard BERT model on Subtask A at OffensEval-2020 \cite{zampieri2020semeval2020}, achieving a score within 2 percentage points of the top performing team. Similarly, \citet{prakash2020incorporating} employed an ensemble model of RoBERTa \cite{liu2019roberta} with a multilayer perceptron using TF-IDF features as input. The ensemble model improved upon the base RoBERTa model by 7 percentage points to achieve state-of-the-art results on the RumourEval-2019 dataset \cite{gorrell-etal-2019-semeval}. We use these studies as a basis for our first hypothesis described in Section \ref{sec:introduction-and-motivation}, and employ a similar method for incorporating TF-IDF features described in Section \ref{sec:methodology}.

Domain-specific information has been shown to be an effective measure of toxicity. \citet{noever2018machine} evaluated the relative predictive value of 28 features of syntax, sentiment, emotion, and outlier word dictionaries for online toxicity detection. By rank-ordering features through feature selection, the most predictive feature of offensive commentary was shown to be a simple bad word list. \citet{pedersen-2019-duluth} compared two logistic regression classifiers against a simple word list model on Subtask A at OffensEval \cite{zampieri2019semeval2019}, with the rule-based model performing 4 percentage points better than either logistic regression model. Section \ref{sec:methodology} discusses our methodology for adding word list features, which capture key word information in the offensive language domain, to pre-trained models.

As discussed in Section \ref{sec:introduction-and-motivation}, a CRF is a method for simultaneous token classification which is not commonly employed by pre-trained models. \citet{souza2020portuguese} proposed a BERT-CRF model architecture composed of a token-level classifier on top of a BERT model followed by a linear-chain CRF. Models with a CRF improved upon or performed similarly to models without one on NER tasks in the Portuguese language. This study is, to the best of our knowledge, one of the few that directly compares the performance of a base BERT model against a BERT-CRF model. The improvements arising from adding a CRF supports our third hypothesis described in Section \ref{sec:introduction-and-motivation}, which aims to explore whether similar improvements will arise in the context of toxic span detection.

Submissions to past toxicity detection tasks at SemEval, such as OffensEval and OffensEval-2020, highlight how effective BERT can be for toxicity detection. \citet{liu-etal-2019-nuli} used a fine-tuned BERT model to achieve state-of-the-art results on Subtask A at OffensEval, and seven of the top ten teams used BERT. Similarly, the top ten teams on Subtask A at OffensEval-2020 all used BERT, RoBERTa, or XLM-RoBERTa \cite{conneau2020unsupervised}, sometimes as part of ensemble models with CNNs and LSTMs. \cite{wiedemann2020uhhlt} submitted the best performing model, which used an ensemble of ALBERT \cite{lan2020albert} models of different sizes. The success of these models in toxicity detection tasks led us to choose to use a BERT-based model for this work.

\section{Methodology}
\label{sec:methodology}

Our pre-trained model of choice was DistilBERT \cite{sanh2020distilbert}, which we used as a baseline measure of performance. We explore and present four models in addition to the baseline DistilBERT model in this paper:

\newpage

\begin{enumerate}[noitemsep,nolistsep]
    \item DistilBERT+TF-IDF
    \item DistilBERT+Word List
    \item DistilBERT+TF-IDF+Word List
    \item DistilBERT+CRF
\end{enumerate}

To build a basis for comparison, all models were trained using the training data provided by the task organisers and evaluated against the validation dataset. The best performing models were then submitted for evaluation against the test dataset during the task evaluation period. The training process was performed five times, using a different random seed each time. This is because varying the random seed used in fine-tuning BERT models can yield substantially different results, even if the models are the same and identical hyperparameters are used \cite{dodge2020finetuning}. The best performing version of each model was used for the remainder of the study.

In Section \ref{sec:introduction-and-motivation}, we hypothesised that adding count-based information to pre-trained models would improve model performance in token classification tasks. TF-IDF is a count-based statistical measure that captures corpus-level information, accounting for global correlations and associations between words. Use of TF-IDF captures word importance, enabling the identification of key words. This word importance could contribute to the identification of a text's toxicity, as shown by \citet{lim2020uob,prakash2020incorporating}. Thus, we tested our first hypothesis by integrating TF-IDF with the DistilBERT model.

One of the most straightforward approaches for toxicity detection is to use a word list, whereby the toxicity of a sequence is determined by comparing the words it contains against a list of known toxic words. Such domain-specific information has been shown to be effective for toxicity detection \cite{noever2018machine,pedersen-2019-duluth}. We tested our second hypothesis by adapting a word list feature for token classification and integrating it with the DistilBERT model.

We incorporated the TF-IDF and word list features by modifying the DistilBERT model. First, we removed the token classification layer on top of the baseline DistilBERT model. Then, for the TF-IDF feature, we appended each token's TF-IDF weight to its hidden state output vector. For the word list feature, we appended a value of 0 or 1. A value of 1 was used if the token appeared in the word list, whilst a value of 0 was used if it did not. These vectors were then pushed through a fully connected layer for classification.

We tested our third hypothesis by adding a CRF, a method for simultaneous token classification, to the DistilBERT model. We followed the more successful fine-tuning approach used by \citet{souza2020portuguese}, which uses a linear classification layer and updates all weights, including BERT's, during training. The CRF takes the output scores from the classification layer as input and computes the log-likelihood of the given sequence of tags. The model was trained to maximise the log-likelihood of the correct tag sequence.

\section{Results}
\label{sec:results}

Table \ref{tab:val-results} shows how the best performing version of each model performed when tested against the validation dataset.

\begin{table}[H]
    \centering
    \begin{tabular}{lr}
        \hline
        Model & F1 Score\\
        \hline
        DistilBERT & 0.58896\\
        DistilBERT+TF-IDF & 0.58930\\
        DistilBERT+Word List & 0.59296\\
        DistilBERT+TF-IDF+Word List & 0.58613\\
        DistilBERT+CRF & 0.58615\\
        \hline
    \end{tabular}
    \caption{Best F1 score achieved by each model, tested against the validation dataset.}
    \label{tab:val-results}
\end{table}

We observe that the inclusion of count-based features did improve model performance, though the increase was very slight. A larger improvement resulted from the use of a word list, whilst model performance worsened when both TF-IDF and word list features were used together and when a CRF was added.

As model performance on the validation dataset was very similar, all models were submitted to the task evaluation stage. Table \ref{tab:test-results} shows the results of each model tested against the test dataset, whilst Table \ref{tab:ranking} shows how our best performing model ranked out of the 91 participating teams.

\begin{table}[H]
    \centering
    \begin{tabular}{lr}
        \hline
        Model & F1 Score\\
        \hline
        DistilBERT & 0.66937\\
        \textbf{DistilBERT+TF-IDF} & \textbf{0.67609}\\
        DistilBERT+Word List & 0.67136\\
        DistilBERT+TF-IDF+Word List & 0.67393\\
        DistilBERT+CRF & 0.67409\\
        \hline
    \end{tabular}
    \caption{F1 score achieved by each model, tested against the test dataset.}
    \label{tab:test-results}
\end{table}

\begin{table}[H]
    \centering
    \begin{tabular}{llr}
        \hline
        Rank & Team & F1 Score\\
        \hline
        1 & HITSZ-HLT & 0.70830\\
        2 & S-NLP & 0.70770\\
        3 & hitmi\&t & 0.69848\\
         & \multicolumn{1}{c}{$\cdots$} & \\
        \textbf{25} & \textbf{UoB} & \textbf{0.67609}\\
        \hline
    \end{tabular}
    \caption{Ranking and F1 score achieved by each team's best performing model, tested against the test dataset.}
    \label{tab:ranking}
\end{table}

It is clear that the F1 scores achieved by all models were very similar when tested against the test dataset, with the DistilBERT+TF-IDF model improving upon the baseline DistilBERT model the most. It is worth noting, however, that whilst this difference is only 0.00672, that same difference would have increased our ranking by 6 ranks had it been added to our final F1 score. Section \ref{sec:discussion-and-analysis} analyses the significance of the results achieved and studies the differences between the predictions of the DistilBERT and DistilBERT+TF-IDF models in greater detail.

\section{Discussion and Analysis}
\label{sec:discussion-and-analysis}

It is difficult to conclude with any certainty whether the addition of our proposed features improved model performance, as the scores achieved are very similar. Whilst the increase in performance observed may indeed be due to our additions to the model, we also propose two alternative theories.

The similarity in results may be due to the relative length of the token vectors compared to the length of the additional features. The hidden output from DistilBERT represents each token as a vector of length 768; the addition of one or two elements to each token vector may not be significant enough to discernibly impact model predictions. That being said, there are still small variations in performance between the models. Whilst this may be due to the addition of new features, it may also be due to variations in the random seed used during fine-tuning. Our most improved model performed 0.00672 F1 points better than the baseline DistilBERT model - a figure within the performance variation range observed during tests involving the random seed (the DistilBERT+TF-IDF model ranged by 0.00922 from 0.58008 to 0.58930, for example). Despite our efforts to counteract this, time and resource limitations meant we were only able to train each model five times instead of the more rigorous ten.

We conduct a more detailed analysis into the predictions of the baseline DistilBERT model and our best performing (DistilBERT+TF-IDF) model. Tables \ref{tab:distilbert-confusion-matrix} and \ref{tab:distilbert-tfidf-confusion-matrix} show the confusion matrix of each model, respectively. These matrices are constructed using a subset of the test dataset from which the  tokens correctly predicted by both models to be non-toxic have been removed. We subset the dataset in this way to significantly reduce the size of the data and to remove less interesting tokens. Table \ref{tab:distilbert-confusion-matrix} shows that the baseline DistilBERT model tends to overpredict toxic labels, resulting in 1466 false positives. Table \ref{tab:distilbert-tfidf-confusion-matrix} shows that the addition of the TF-IDF feature helps to mitigate this, with the DistilBERT+TF-IDF model correctly predicting over 100 of DistilBERT's false positives as true negatives - an overall improvement of 2.5\% on this subset of the test set.

\begin{table}[H]
    \centering
    \begin{tabular}{cc|c|c|}
         & \multicolumn{1}{c}{} & \multicolumn{2}{c}{\textbf{Predicted}}\\
         & \multicolumn{1}{c}{} & \multicolumn{1}{c}{\textbf{Non-Toxic}} & \multicolumn{1}{c}{\textbf{Toxic}}\\
        \cline{3-4}
        \multicolumn{1}{c}{\multirow{2}{*}{\rotatebox{90}{\textbf{True}}}}
         & \textbf{Non-Toxic} & 202 & 1466\\
        \cline{3-4}
         & \textbf{Toxic} & 682 & 1829\\
        \cline{3-4}
    \end{tabular}
    \caption{Confusion matrix of the DistilBERT model on a subset of the test dataset from which the tokens correctly predicted by both the DistilBERT and DistilBERT+TF-IDF models to be non-toxic have been removed.}
    \label{tab:distilbert-confusion-matrix}
\end{table}

\begin{table}[H]
    \centering
    \begin{tabular}{cc|c|c|}
         & \multicolumn{1}{c}{} & \multicolumn{2}{c}{\textbf{Predicted}}\\
         & \multicolumn{1}{c}{} & \multicolumn{1}{c}{\textbf{Non-Toxic}} & \multicolumn{1}{c}{\textbf{Toxic}}\\
        \cline{3-4}
        \multicolumn{1}{c}{\multirow{2}{*}{\rotatebox{90}{\textbf{True}}}}
         & \textbf{Non-Toxic} & 310 & 1358\\
        \cline{3-4}
         & \textbf{Toxic} & 692 & 1819\\
        \cline{3-4}
    \end{tabular}
    \caption{Confusion matrix of the DistilBERT+TF-IDF model on a subset of the test dataset from which the tokens correctly predicted by both the DistilBERT and DistilBERT+TF-IDF models to be non-toxic have been removed.}
    \label{tab:distilbert-tfidf-confusion-matrix}
\end{table}

In addition to an exploration of the results and confusion matrix, we perform an error analysis by manually comparing the true labels and predictions of the DistilBERT and DistilBERT+TF-IDF models. We first observe that there are some inconsistencies in the true labels. For example, the phrase ``\dots racist, sexist, narcissistic, pathological liar \dots'' is marked as non-toxic, whereas ``\dots sexist rubbish \dots'' is marked as toxic. Another trend observed is that both models struggle to correctly predict phrases containing ordinarily non-toxic words which become toxic given the context. For example, consider the phrases ``Trump troll'', ``Bunch of cowards'', ``Total rubbish'', and ``PATHETIC LIB LOSER''. Whilst the true labels classify all of these phrases as toxic, both models only predicted the words ``troll'', ``cowards'', ``rubbish'', ``pathetic'', and ``loser'' to be toxic.

These trends highlight some of the inherent difficulties involved in token classification tasks for both machine learning models and human annotators.

\section{Conclusion and Future Work}
\label{sec:conclusion}

This work explored the possibility of improving pre-trained model performance on the token classification task of toxic span detection. As discussed in Section \ref{sec:introduction-and-motivation}, we hypothesised that adding 1) count-based information, 2) domain-specific knowledge, and 3) a CRF can aid pre-trained models in token classification tasks. Whilst our experimental results (Section \ref{sec:results}) seem to suggest that all three of these features improve the performance of DistilBERT, we note that they do so only marginally (Section \ref{sec:discussion-and-analysis}). Further analysis, however, showed that, whilst the overall F1 improvement from adding TF-IDF was small, the addition of a count-based feature helped to reduce DistilBERT's overprediction of toxic tokens.

We believe that these improvements, whilst small, provide an interesting avenue of exploration. We intend to further explore how these and other similar features interact with pre-trained models in the task of token classification.



\bibliographystyle{acl_natbib}
\bibliography{acl2021}


\end{document}